\documentclass{article}

\usepackage[nonatbib, final]{nips_2018}

\usepackage[utf8]{inputenc} 
\usepackage[T1]{fontenc}    
\usepackage{hyperref}       
\usepackage{url}            
\usepackage{booktabs}       
\usepackage{amsfonts}       
\usepackage{nicefrac}       
\usepackage{microtype}      
\usepackage{graphicx}
\usepackage{multirow}
\usepackage{float}
\usepackage{caption}

\title{Integrating Reinforcement Learning to Self Training for Pulmonary Nodule Segmentation in Chest X-rays}

\author{
  Sejin Park\thanks{Equal Contribution} \\
  VUNO Inc.\\
  Seoul, South Korea\\
  \texttt{gnoses@gmail.com} \\
  \And
  Woochan Hwang\footnotemark[1] \\
  Imperial College London \\
  United Kingdom \\
  \texttt{wh1714@ic.ac.uk} \\
  \And
  Kyu-Hwan Jung \\
  VUNO Inc.\\
  Seoul, South Korea \\
  \texttt{khwan.jung@vuno.co} \\
}

\begin{document}

\maketitle

\begin{abstract}
 
Machine learning applications in medical imaging are frequently limited by the lack of quality labeled data. In this paper, we explore the self training method, a form of semi-supervised learning, to address the labeling burden. By integrating reinforcement learning, we were able to expand the application of self training to complex segmentation networks without any further human annotation. The proposed approach, {\it reinforced self training (ReST)}, fine tunes a semantic segmentation networks by introducing a policy network that learns to generate pseudolabels. We incorporate an expert demonstration network, based on inverse reinforcement learning, to enhance clinical validity and convergence of the policy network. The model was tested on a pulmonary nodule segmentation task in chest X-rays and achieved the performance of a standard U-Net while using only 50\% of the labeled data, by exploiting unlabeled data. When the same number of labeled data was used, a moderate to significant cross validation accuracy improvement was achieved depending on the absolute number of labels used. 
\end{abstract}

\section{Introduction}
Supervised learning applications in medical imaging face a common obstacle of obtaining quality labeled data. This problem is particularly an issue for segmentation tasks due to the lack of standardized annotation practices for many segmentation labels \cite{kohli2017medical} and the inherent ambiguity in radiology, shown by a retrospective 20-year literature review \cite{goddard2001error, muhm1983lung}. Even with access to radiologists who can review or annotate new datasets, it is immensely time consuming and expensive to scale to the amount where supervised learning algorithms can be implemented effectively. This confines most applications to problems with a clear financial return while neglecting many with a potentially huge clinical impact.

One way of overcoming such limitations is through semi-supervised learning \cite{Zhu2009}, which aims to exploit the relatively abundant unlabeled images. There are many forms of semi-supervised learning including, but not limited to, self training \cite{mcclosky2006effective}, generative methods \cite{kingma2014semi}, co-training \cite{blum1998combining} and active learning \cite{Settles2012}. Among these, we will focus on the self training method, which can easily be incorporated to existing complex neural networks. A basic self training method will employ thresholding on the model output space to generate pseudolabels \cite{lee2013pseudo} that are used to fine tune an existing model. However, such simple decision policies generate noisy pseudolabels, resulting in non-convergence when applied to semantic segmentation networks. The few published work on self training methods for segmentation tasks involve limitations such as mannually generated regions of interest \cite{azmi2011impst} or only using negative pseudolabels \cite{park2018false}. 

In this work we propose reinforced self training (ReST), a more sophisticated approach to pseudolabel generation, where we integrate reinforcement learning to train a policy network that generates pseudolabels to maximize the validation accuracy. Unlike conventional self training methods, which exploits the assumption of low density separation between classes \cite{chapelle2005semi}, this model finds optimal pseudolabels by exploring with feedback from the changing validation accuracy. However, the validation accuracy alone is a very sparse reward that leads to a challenging optimization problem that many reinforcement learning approaches face. To tackle this issue, we train a classifier with expert demonstrations to determine if the generated labels are sufficiently similar to that of the experts' and use the classifier output as an auxiliary reward. The integration of the classifier in the reward definition is crucial for stable convergence as it provides a dense reward function and prevents early mistakes, which can reinforce themselves. Furthermore, the classifier reward acts as a penalty term against generating clinically nonsensical labels, making our framework suitable for solving problems in the medical imaging domain. We will demonstrate the potential of our approach on the pulmonary nodule segmentation task in chest X-rays.

To the best of our knowledge, this work is the first integration of reinforcement learning and self training for image segmentation networks. We propose a new approach to decreasing the labeling burden and a model design for reinforcement learning that prevents convergence to a clinically invalid hypothesis.

\section{Methods}

\subsection{Data Collection and Preprocessing}

The chest X-rays (PA view) used were collected between 2013 and 2015 from Seoul Asan Medical Center. The dataset is comprised of 931 images with pixel labels (1007 pulmonary nodules) and 2986 images without labels. The labels were created by consensus from three board certified radiologists with 10 to 25 years of experience. 
Apart from per image histogram equalization to mitigate intensity variation, no further augmentations were applied.

\subsection{Proposed Model: ReST}

In the proposed model we begin with a segmentation network (environment) which is trained in the conventional supervised learning approach. Then we implement ReST as a wrapping method to further improve the performance of the supervised segmentation network. A policy network \cite{sutton2000policy} (agent) learns to generate pseudolabels for the unlabeled data based on the validation accuracy of the segmentation network. To aid the sparse reward and prevent the agent generating clinically nonsensical labels, we use inverse reinforcement learning (IRL) \cite{Abbeel2004} to train a classifier that will decide if the generated pseudolabels are sufficiently similar to that of the experts'. The trained classifier is used to provide initial rewards to the policy network during the exploration stage.

\begin{table}[htbp]
\centering
\begin{tabular}{l}
\hline\hline
\textbf{Algorithm 1} Reinforced Self Training \hspace*{80mm}
 \\ \hline 
  \rule{0pt}{2ex}\textbf{Input: } Unlabeled set $U$, labeled set $L$ \\
  \textbf{Output: } Optimal policy $\pi^{*}$ \\
  0 : Train segmentation network $S \gets L \hfill \triangleright$ supervised learning \\
  1 : Train expert reward function R$^{*}_{exp} \gets \{S(L), L \}  \hfill \triangleright$ max margin planning\\
  2 : Initialize policy network $\pi_{0}$ \\
  3 : \textbf{for} iteration 1:k \textbf{do} $\hfill \triangleright$ self training\\
  4 : $\;\;\;\;$Generate pseudolabel set $P = \pi(U)$   $\hfill \triangleright$ exploration \\
  5 : $\;\;\;\;$Update $\pi \gets$ R$^{*}_{exp}(P)$ \\
  6 : $\;\;\;\;$\textbf{if } R$^{*}_{exp}(P) > $ threshold \textbf{do} \\ 
  7 : $\;\;\;\;\;\;\;\;$Update $S \gets $ P and collect validation accuracy R$_{val} \hfill \triangleright$ exploitation\\
  8 : $\;\;\;\;\;\;\;\;$Update $\pi \gets$ R$_{val}$ \\
  9 : \textbf{end for}
 \\ \hline
\end{tabular}
\end{table}

\subsubsection{Environment Set Up}

\textbf{Supervised Learning:} A U-Net \cite{Ronneberger2015} like segmentation model is trained in four different settings, each using 25\%, 50\%, 75\% and 100\% of the available labeled data. These will be compared after fine tuning with the proposed method to prove its efficacy in reducing the labeling burden. The trained model becomes the environment that interacts with the agent in the following description. 

\textbf{Expert Reward Function:} After initial convergence of the environment network, a set of expert demonstrations is created using the segmentation network output of the labeled training data and its true labels generated from board certified radiologists. The expert reward function $R^{*}_{exp}$, a binary classification network, is approximated from the experts behavior via maximum margin IRL \cite{Ratliff2006}. The same label set used for the supervised learning stage is used for training and no iterative updates are performed on the reward function after this stage.

The expert reward function acts as a virtual radiologist that decides if the action taken by the agent (i.e. the generated label) in a given state is sufficiently similar to that of the board certified radiologists. The network will give a value of 1 if sufficiently similar and 0 otherwise. 

\subsubsection{Agent Training}

Let a Markov Decision Process (MDP) be defined as $M$ = \{$S$, $A$, $T$, $R$\}, where $S$ denotes the state space, $A$ denotes the action space, $T$ is the transition dynamics and $R$ is the reward. In the given active learning setting, the state space of the MDP is the output of the trained U-net segmentation model and the action is to find pseudolabels on unlabeled data. The transition dynamics is defined by the stochastic policy network $\pi$. We define the reward R as either the output of the IRL network $R^{*}_{exp}$ or the in-training validation accuracy $R_{val}$ of the U-net depending on the stage of training. 

\textbf{Initialization:} As with many reinforcement learning algorithms, we introduce an initial heuristic method to support the unstable policy $\pi$ during the early stages of agent training. A sample is considered informative if the entire state space is below the negative threshold, in which case will be labeled normal, or contain regions above the positive threshold, where the region of interest will be considered a true nodule. Additionally, connected components labeling was used to filter out regions that are above the positive threshold but smaller than 0.1 mm as it is unlikely to be a true nodule. We exploit the $e$-greedy exploration method to choose between the stochastic policy of the RL agent and the heuristic method. 

\textbf{Training:} The agent performs a pixel level binary classification to decide which pixels are part of a pulmonary nodule based on the current policy. The generated pseudolabel set is evaluated by the expert classifier $R_{exp}^{*}$, and the average score of the pseudolabel set is used as a reward to update the policy $\pi$ according to its gradient. When the average score of the expert reward function surpasses a predefined threshold, we lower the randomness of the stochastic policy $\pi$ and begin to use the pseudolabels to fine tune our segmentation network. In this stage, the validation accuracy of the segmentation network $R_{val}$ is used as a reward to update policy $\pi$, instead of $R_{exp}^{*}$. Once the segmentation network performance stabilizes, we return to the exploration stage using $R_{exp}^{*}$. The use of the expert classifier as an auxiliary reward function is crucial as the sparsity of $R_{val}$ leads to unstable training.

\section{Results}

In our evaluation, we compare the F1 score of a standard U-Net segmentation network before and after fine tuning via the proposed approach. We also compare the results when the environment is initialized with different number of labeled data in the supervised learning phase before applying ReST. Equal number of unlabeled data is used for all cases.

\begin{table}[htbp]
\small
\centering
\caption{Validation results (five repeats of five fold validation) of the proposed framework initialized with different proportions of the available labeled data. The t-Tests (given as p-values) compare F1 scores of a standard U-Net segmentation network to the F1 score after fine tuning with ReST.\\}
\label{my-label}
\begin{tabular}{l|lllllll}
Labeled & \multicolumn{3}{c|}{Standard U-Net}                                                 & \multicolumn{3}{c|}{ReST}                                                 & t-Test \\
data used & \multicolumn{1}{c|}{F1} & \multicolumn{1}{c|}{Sensitivity} & \multicolumn{1}{c|}{FPs/Img} & \multicolumn{1}{c|}{F1} & \multicolumn{1}{c|}{Sensitivity} & \multicolumn{1}{c|}{FPs/Img} & (F1) \\ \hline

25\%                               & 0.738 $\pm$ 0.015                                                       & 0.732                            & 0.675                         & 0.764 $\pm$ 0.027                           & 0.780                            & 0.742                         & \textless 0.001                                    \\ \hline

50\%                               &  0.745 $\pm$ 0.018                                                      & 0.772                            & 0.508                         & 0.802 $\pm$ 0.014                           & 0.829                            & 0.312                         & \textless 0.001                                    \\ \hline

75\%                               & 0.794 $\pm$ 0.023                                                      & 0.822                            & 0.534                         & 0.821 $\pm$ 0.019                           & 0.865                            & 0.262                         & \textless 0.001                                    \\ \hline

100\%                              & 0.812 $\pm$ 0.014                                                      & 0.856                            & 0.342                         & 0.848 $\pm$ 0.022                           & 0.887                            & 0.252                         & \textless 0.001                                    \\ \hline
\end{tabular}
\end{table}
\setlength{\floatsep}{4pt plus 1.0pt minus 2.0pt}

The average of five repeats of five fold validation performance before and after applying ReST to a segmentation network trained with different proportions of the available labeled data is shown in table 1. The p-value indicates the results of the t-Test performed on the F1 score before and after applying ReST for each setting. The result demonstrates that the ReST approach improves the cross validation accuracy of the segmentation model throughout. 

The standard U-Net performance achieved using 100\% of the labeled data is the best performance of a segmentation network under supervised learning with the given labeled data. Table 1 shows that with the ReST approach, we can achieve the same level of performance (F1 score) with only 50\% of the labeled data (p-value < 0.05). Figure 1 shows the performance convergence of ReST relative to standard self training \cite{mcclosky2006effective} and pseudolabel mining \cite{park2018false}. An example inference result on the validation set can be seen in figure 2. 

\begin{figure}[htbp]
\centering
\includegraphics[width=90mm]{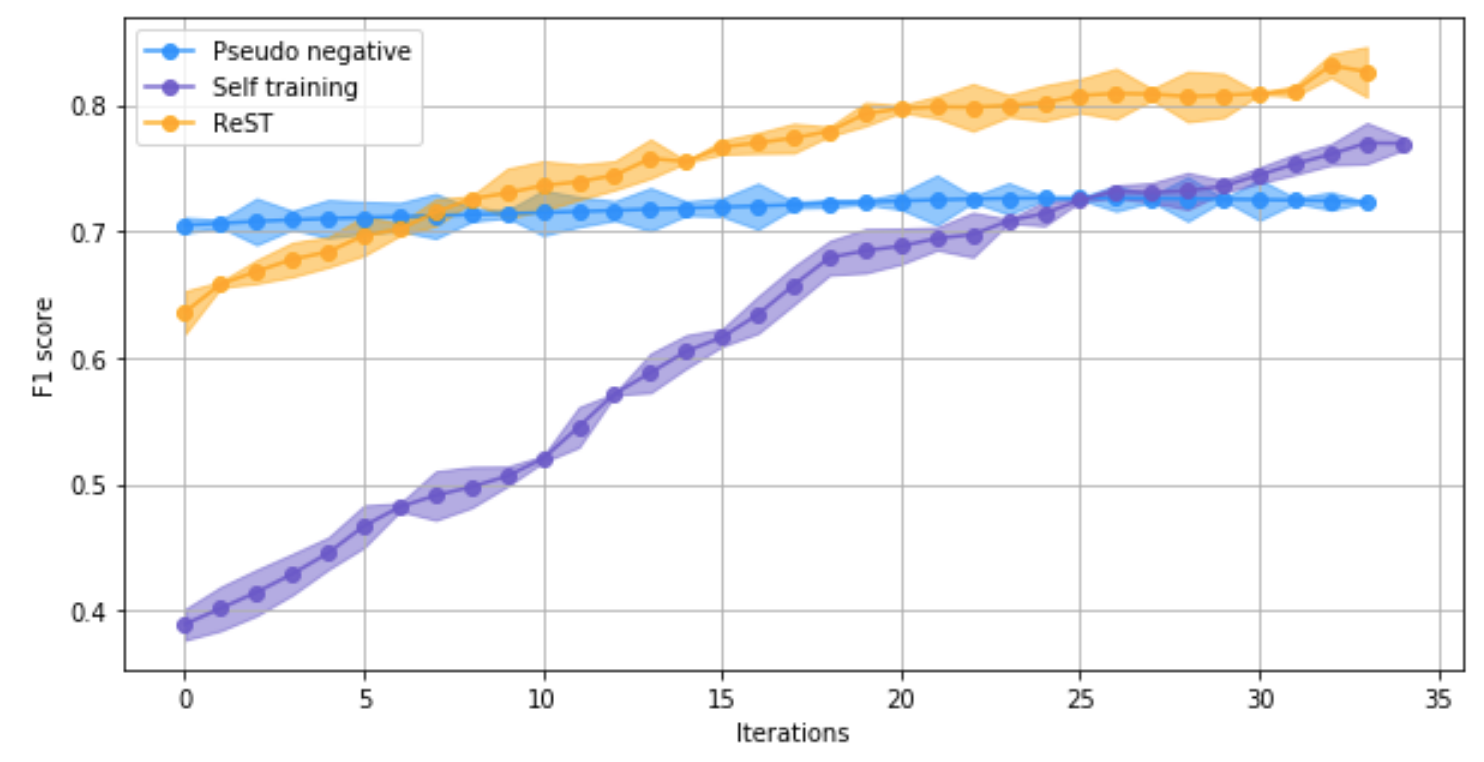}
\caption{Performance over training iterations of the proposed framework (initialized with 75\% of the labeled data) compared to standard self training \cite{mcclosky2006effective} and the pseudonegative mining method \cite{park2018false}.}
\end{figure}


\begin{figure}[htbp]
\centering
\includegraphics[width=100mm]{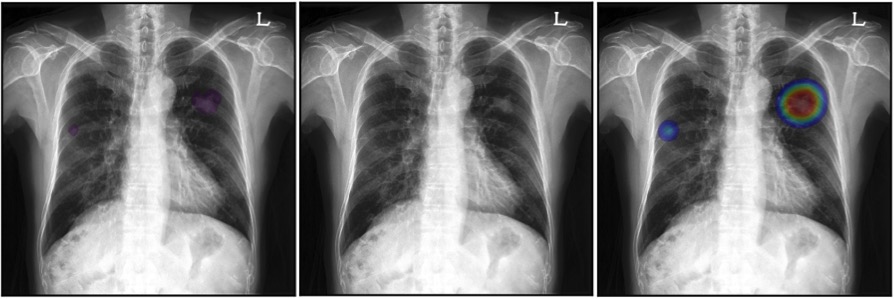}
\caption{Inference result of the model (initialized with 50\% of the labeled data) on validation data. \ (Left) Ground truth, (Middle) Standard U-Net, (Right) Post-ReST}
\end{figure}

\section{Conclusion and Future Work}

We presented a form of self training method that allows a complex contextual approach to pseudolabel generation using deep reinforcement learning. This can be viewed in the context of balancing exploration and exploitation. While standard self training is focused on exploiting the assumption of low density separation between classes, our approach allows more exploration guided by our reward function. Although this inevitably leads to increased computational cost, we were able to successfully demonstrate application of self training to medical image segmentation with no further expert annotations for the first time. 

Our evaluation on the pulmonary nodule detection task in chest X-rays using the U-Net segmentation network showed that our approach can effectively leverage unlabeled data to improve performance of deep neural networks. In the specific dataset and task, we were able to reduce the labeling burden to 50\% while maintaining performance. Though further experimentation on different datasets and tasks is necessary to gauge the true value of this approach, our results show great potential. A particularly interesting area where this framework may be useful is the multi-center adaptation problem. With some consideration in adjusting the reward function, our approach could be used to generalized a pretrained algorithm to settings with different patient demographics without any further labeled data. 

\small
\section*{Acknowledgements}
This study was supported by the Industrial Strategic Technology Development Program of the Ministry of Trade, Industry \& Energy (10072064) in the Republic of Korea.

\small
\bibliography{ReST}

\end{document}